\documentclass[journal,onecolumn,12pt]{IEEEtran}
\usepackage{amsmath}
\usepackage{graphicx}
\usepackage[table,xcdraw]{xcolor}
\usepackage{epsfig}
\usepackage{xcolor}
\usepackage{amssymb}
\usepackage{mathrsfs}

\usepackage{hyperref}
\hypersetup{
	linkcolor  = red,
	citecolor  = green,
	urlcolor   = blue,
	colorlinks = true,
}

\begin{document}

\title{Is Supervised Learning With Adversarial Features Provably Better Than Sole Supervision?}

\author{Litu~Rout$\dagger$ \thanks{$\dagger$Optical Data Processing Division, Signal and Image Processing Group, SAC, ISRO, Ahmedabad, India - 380015. mail id: lr@sac.isro.gov.in}}
	
\maketitle

\begin{abstract}
Generative Adversarial Networks (GAN) have shown promising results on a wide variety of complex tasks. Recent experiments show adversarial training provides useful gradients to the generator that helps attain better performance. In this paper, we intend to theoretically analyze whether supervised learning with adversarial features can outperform sole supervision, or not. First, we show that supervised learning without adversarial features suffer from vanishing gradient issue in near optimal region. Second, we analyze how adversarial learning augmented with supervised signal mitigates this vanishing gradient issue. Finally, we prove our main result that shows supervised learning with adversarial features can be better than sole supervision (under some mild assumptions). We support our main result on two fronts (i) expected empirical risk and (ii) rate of convergence.
\end{abstract}


\section{Introduction}
\label{intro}
Over the past few years, the advancement of deep neural networks has opened up unprecedented opportunities in complex real world problems. The advent of high end computing infrastructure has played a vital role in this remarkable progress. Of particular interest, supervised learning, a domain of artificial intelligence focusing on learning via paired supervised training samples, has been quite effective in wide variety of challenging problems~\cite{huang2017densely, badrinarayanan2017segnet, wang2019deep}.

Despite the progress, it is often useful to heed the difficulty in acquiring sufficient amount of paired data for reliable supervised training. In this regard, the discovery of Generative Adversarial Networks (GANs) has provided a mechanism to reduce human effort in preparation of training data. By bringing this insight into fruition, computer vision problems, where it is almost impossible to gather paired data, are now being addressed with fair amount of certainty~\cite{goodfellow2014generative, isola2017image, zhu2017unpaired}. In particular, the requirement of paired supervised training samples is relaxed to some extent due to the vantage of GANs. The adversarial game between generator and discriminator allows generation of realistic looking artificial samples. Particularly intriguing is the phenomenon of generating samples from a high dimensional distribution without even explicitly estimating its density. From this point of view, an adversarial generator succinctly learns to generate realistic looking samples lying on a compact manifold of low dimensional space.

In recent years, GANs are being used in addressing problems which were believed to be extremely challenging. The pervasive use of GANs has drawn a significant attention of the research community in various domain. Among many applications, some require that a particular sample is generated subject to conditional inputs. For this reason, recent methods propose to regularize the generation process through expert feedback. In photo-realistic image super resolution, the empirical risk of generator is regularized by a metric that minimizes distance between predicted and actual high resolution image~\cite{ledig2017photo}. In visual object tracking via adversarial learning, Euclidean norm is used to regulate the generated mask such that it lies within a small neighbourhood of the actual mask identifying discriminative features~\cite{song2018vital}. In medical image segmentation, multi-scale $L_1$-loss with adversarial features is shown to achieve better performance in terms of state-of-the-art evaluation metrics~\cite{xue2018segan}. Performance gain in these diverse practical applications provides a clear indication of better empirical results of adversarial learning.

Numerous attempts seek to provide empirical evidence on generative adversarial networks outperforming previously used sole supervised approaches. Recent studies suggest purely supervised learning driven reconstructed images have inferior visual perceptual quality as compared to adversarial learning~\cite{wang2019deep}. So far the theoretical investigation shows that the empirical risk of supervised learning augmented with adversarial features does not become arbitrarily large during training. Hence, there exists a small constant that bounds the total empirical risk above~\cite{xue2018segan}. However, these benign properties of loss surface does not necessarily provide enough theoretical evidence on whether supervised learning augmented with adversarial features is better than sole supervision, or not. Furthermore, the regularized generator achieves faster convergence due to efficient flow of useful gradients from the discriminator, but the theoretical understanding remains elusive. Therefore, several questions arise:
\begin{center}
\begin{itemize}
    \item Why do updates take longer time to converge in case of completely supervised learning as compared to regularized adversarial learning?
    
    \item Do adversarial features alleviate this slow convergence issue?
    
    \item Is supervised learning augmented with adversarial features provably better than sole supervision?
    
    \item If so, under which circumstance and on what basis it is better?
\end{itemize}    
\end{center}

\subsection{Summary of Contributions}
\label{summ}
The fundamental contributions of this paper are the answers to these aforementioned questions. Specifically, we provide theoretical evidence to corroborate our answers. It is to be noted that we interchangeably use supervised learning with adversarial features and adversarial learning with expert regularization. By expert regularization we directly minimize a distance between predicted and true samples.

\begin{itemize}
    \item We show that a purely supervised objective suffers from vanishing gradient issue within the tiny landscape of empirical risk, provided the trainable parameters fall in the near optimal region.
    
    \item Further, we provide mathematical explanations on adversarial discriminator being able to mitigate the issue of vanishing gradient under some mild assumptions.
    
    \item As a part of our main contribution, we finally prove that supervised learning with adversarial features can be provably better than a purely supervised learning task. 
    
    \item More broadly, our theoretical investigation suggests that by augmenting adversarial features in a supervised learning framework, the expected empirical risk and rate of convergence is guaranteed to be at least as good as sole supervision.
\end{itemize}

\section{Preliminaries}
\label{prelim}
Here, we briefly explain the architectures under study and summarize our notations. Given positive integers $a$ and $b$, where $a < b$, by $\left [ a \right ]$ we mean the set $\left \{1,2,\dots,a \right \}$, and by $\left [a,b \right ]$, the set $\left \{a,a+1,\dots,b \right \}$. Let $X \subset \mathbb{R}^{d_{x}}$, $Y \subset \mathbb{R}^{d_{y}}$, and $Z \subset \mathbb{R}^{d_{z}}$. Given a vector $x$, $\left \| x \right \|$ represent the Euclidean norm. Given a matrix $M$, $\left \| M \right \|$ represent the spectral norm. By $f(\theta)|_{\theta_i}$, we mean $f(\theta)$ evaluated at $\theta_i$.

Let $x \in X$ be the input vector. We consider an $L$-block resnet, $f_{\theta}(.)$ as the common architecture for generator of adversarial network and supervised learning. The output is computed as following,
\begin{equation*}
    \begin{split}
        f_{\theta}(x) & = \omega^{\textit{T}} h_L(x), \\
        h_l(x) & = h_{l-1}(x) + V_l\phi_z^{l}(U_l h_{l-1}(x)), l=1,2,\dots,L, \\
        h_0 (x) & = x.
\end{split}
\end{equation*}
Here,  $\phi_{z} (.)$ represents a neural network with parameters $z$. $\theta$ denotes the collection of parameters \\ $\left \{w,z,U_1,V_1,U_2,V_2,\dots,U_L,V_L \right \}$ of appropriate dimensions~\cite{yun2019deep}. The discriminator, $g_{\psi}(.)$ of adversarial network has trainable parameters collected by $\psi$. By $\mathcal{J}_{\theta}\left ( f_{\theta}\left ( x \right ) \right )$, we mean Jacobian matrix of $f_\theta(x)$ evaluated at $\theta$.

\section{Motivation}
\label{motiv}
\subsection{Vanishing Gradient of Supervised Objective in Near Optimal Region}
\label{th1}
\textbf{Assumption 3.1} The loss function $l(p;y)$ is a convex and continuously differentiable function of $p$, i.e., $l(p;y) \in {\mathscr{C}}'(Y)$. We also assume $l(p;y)$ be a locally $K-$Lipschitz, i.e., given $y \in Y$, $|{l}'(p;y)| \leq K,~\forall~p$.

\textbf{Theorem 3.1} \textit{Suppose \textbf{Assumption 3.1} holds. Let $f_\theta: X \mapsto Y $ be a differentiable function.  Let $\mathcal{P}$ be an empirical distribution over training samples. If (i) $\mathbb{E}_{(x,y)\sim \mathcal{P}}\left [ \left \| \mathcal{J}_{\theta}\left ( f_{\theta}\left ( x \right ) \right ) \right \|^{2} \right ]\leq M^2$ and (ii) trainable parameters are in the near optimal region, i.e., $\mathbb{E}_{(x,y)\sim \mathcal{P}}\left [ \left \| f_\theta(x) - f_{\theta^*}(x) \right \| \right ]\leq \epsilon$, then the expected gradient of purely supervised objective vanishes. That is,\\
\begin{center}
    $\left \|\nabla_\theta \mathbb{E}_{(x,y)\sim \mathcal{P}}\left [ l\left ( f_\theta(x);y \right ) \right ] \right \| \leq \lambda M$.
\end{center}
}

\textit{Proof.} 
\begin{equation*}
    \begin{split}
        \left \|\nabla_\theta \mathbb{E}_{(x,y)\sim \mathcal{P}}\left [ l\left ( f_\theta(x);y \right ) \right ] \right \|^2 & \leq \mathbb{E}_{(x,y)\sim \mathcal{P}}\left [ \left \|\nabla_\theta  l\left ( f_\theta(x);y \right )  \right \|^2 \right ] \\
        & \leq \mathbb{E}_{(x,y)\sim \mathcal{P}}\left [ \left \|\nabla_{\hat{y}} l\left ( f_\theta(x);y \right ) \nabla_\theta f_\theta(x)  \right \|^2 \right ], \text{where}~\hat{y} = f_\theta(x)\\
        & \leq \mathbb{E}_{(x,y)\sim \mathcal{P}}\left [ \left \|\nabla_{\hat{y}} l\left ( f_\theta(x);y \right )\right \|^2 \left \|\nabla_\theta f_\theta(x)  \right \|^2 \right ] \\
        & \leq \mathbb{E}_{(x,y)\sim \mathcal{P}}\left [ \left \|\nabla_{\hat{y}} l\left ( f_\theta(x);y \right )\right \|^2\right ] \mathbb{E}_{(x,y)\sim \mathcal{P}}\left [ \left \| \mathcal{J}_{\theta}\left ( f_{\theta}\left ( x \right ) \right ) \right \|^{2} \right ]
\end{split}
\end{equation*}
 By continuously differentiable property, it is required that to every $q \in Y$ and to every $\lambda \geq 0$ corresponds an $\epsilon \geq 0$ such that if $p\in Y$ and $\left \| p-q \right \|\leq \epsilon$, then $\left \| {l}'(p;y) - {l}'(q;y) \right \|\leq \lambda$. Now, substitute $p=f_\theta(x)$ and $q=f_{\theta^*}(x)$. Condition $\mathbb{E}_{(x,y)\sim \mathcal{P}}\left [ \left \| f_\theta(x) - f_{\theta^*}(x) \right \|_2 \right ]\leq \epsilon$ holds. Therefore, 
\begin{equation}
\label{eq31}
    \mathbb{E}_{(x,y)\sim \mathcal{P}}\left [ \left \| {l}'(f_\theta(x);y) - {l}'(f_{\theta^*}(x);y) \right \|\right ] \leq \lambda
\end{equation}

Since $\mathbb{E}_{(x,y)\sim \mathcal{P}}\left [ \left \| {l}'\left ( f_\theta(x);y \right ) - {l}'\left ( f_{\theta^*}(x);y \right ) \right \| \right ] \geq \mathbb{E}_{(x,y)\sim \mathcal{P}}\left [ \left \| {l}'\left ( f_\theta(x);y \right )\right \| \right ] - \mathbb{E}_{(x,y)\sim \mathcal{P}}\left [ \left \| {l}'\left ( f_{\theta^*}(x);y \right )\right \| \right ]$, equation~(\ref{eq31}) implies,
\begin{equation*}
    \begin{split}
         & \mathbb{E}_{(x,y)\sim \mathcal{P}}\left [ \left \| {l}'\left ( f_\theta(x);y \right )\right \| \right ] - \mathbb{E}_{(x,y)\sim \mathcal{P}}\left [ \left \| {l}'\left ( f_{\theta^*}(x);y \right )\right \| \right ]  \leq \lambda \\
     & \implies \mathbb{E}_{(x,y)\sim \mathcal{P}}\left [ \left \| {l}'\left ( f_\theta(x);y \right )\right \| \right ]  \leq  \lambda,~ (\mathbb{E}_{(x,y)\sim \mathcal{P}}\left [ \left \| {l}'\left ( f_{\theta^*}(x);y \right )\right \| \right ] = 0,~\because \theta^* \text{ is optimal})\\
     & \implies \mathbb{E}_{(x,y)\sim \mathcal{P}}\left [ \left \| \nabla_{\hat{y}} l\left ( f_\theta(x);y \right )\right \|^2 \right ] \leq \lambda^2
    \end{split} 
\end{equation*}

Now,
\begin{equation*}
    \begin{split}
        & \left \|\nabla_\theta \mathbb{E}_{(x,y)\sim \mathcal{P}}\left [ l\left ( f_\theta(x);y \right ) \right ] \right \|^2 \leq \lambda^2 M^2 \\
        & \implies \left \|\nabla_\theta \mathbb{E}_{(x,y)\sim \mathcal{P}}\left [ l\left ( f_\theta(x);y \right ) \right ] \right \| \leq \lambda M. \text{ This finishes the proof. \hfill} \square
    \end{split}
\end{equation*}

\textbf{Theorem 3.1} provides an upper bound on the expected gradient over empirical distribution $\mathcal{P}$ in near optimal region. The expected gradient shrinks in proportional to spectral norm of Jacobian matrix and approximation error. Thus, the small gradients in purely supervised learning exacerbates training progress within this tiny landscape of empirical risk. Furthermore, the rate of convergence slows down drastically. In other words, this shows the gradient updates become smaller as the training progresses which resonates with intuitive understanding of gradient descent. Therefore, a fundamental question arises.  Can we attain convergence faster without having to loose any empirical risk benefits? We discuss this question in the following section.

\subsection{Mitigating Vanishing Gradient with Adversarial Features}
\label{th23}
\textbf{Wasserstein Objective:} The generator cost function of WGAN is given by,
\begin{equation*}
   \arg \min_{\theta} -\mathbb{E}_{x\sim \mathcal{P}_X}\left [ g_{\psi}\left ( f_\theta \left ( x \right ) \right ) \right ],
\end{equation*}
and the discriminator cost function,
\begin{equation}
\label{dis}
    \arg \min_{\psi} \mathbb{E}_{x\sim \mathcal{P}_X}\left [ g_{\psi}\left ( f_\theta \left ( x \right ) \right ) \right ] - \mathbb{E}_{y\sim \mathcal{P}_Y}\left [ g_{\psi}\left ( y \right ) \right ].
\end{equation}

\textbf{Theorem 3.2} \textit{Suppose condition $(i)$ of \textbf{Theorem 3.1} holds. Let $g_\psi:Y \mapsto \mathbb{R}$ be a differentiable discriminator. If  $\left \| g-g^* \right \| \leq \delta$, where $g^*:=g_{\psi^*}$ denote optimal discriminator, then \\
\begin{center}
    $\left \|-\nabla_\theta \mathbb{E}_{x\sim \mathcal{P}_X}\left [ g_{\psi}\left ( f_\theta \left ( x \right ) \right ) \right ]\right \| \leq \delta M$.
\end{center}
}
\textit{Proof.}
\begin{equation*}
    \begin{split}
        \left \|-\nabla_\theta \mathbb{E}_{x\sim \mathcal{P}_X}\left [ g_{\psi}\left ( f_\theta \left ( x \right ) \right ) \right ]\right \|^2 & \leq \mathbb{E}_{x\sim \mathcal{P}_X}\left [ \left \|\nabla_\theta g_{\psi}\left ( f_\theta \left ( x \right ) \right )\right \|^2 \right ]\\
        & \leq \mathbb{E}_{x\sim \mathcal{P}_X}\left [ \left \|\nabla_{\hat{y}} g_{\psi}\left ( f_\theta \left ( x \right ) \right )\right \|^2 \left \|\nabla_{\theta}f_\theta(x)) \right \|^2 \right ], \text{ where } \hat{y}=f_\theta(x)\\
        & \leq \mathbb{E}_{x\sim \mathcal{P}_X}\left [ \left \|\nabla_{\hat{y}} g_{\psi}\left ( f_\theta \left ( x \right ) \right )\right \|^2\right ] \mathbb{E}_{x\sim \mathcal{P}_X}\left [ \left \|\nabla_{\theta}f_\theta(x)) \right \|^2 \right ] \\
        & \leq \mathbb{E}_{x\sim \mathcal{P}_X}\left [ \left (\left \|\nabla_{\hat{y}}   g_{\psi^*}\left ( f_\theta \left ( x \right ) \right )\right \| + \delta\right )^2\right ] \mathbb{E}_{(x,y)\sim \mathcal{P}}\left [ \left \| \mathcal{J}_{\theta}\left ( f_{\theta}\left ( x \right ) \right ) \right \|^{2} \right ] \\
        & \leq \delta^2 M^2, \left (\left \|\nabla_{\hat{y}}   g_{\psi^*}\left ( f_\theta \left ( x \right ) \right )\right \|=0,~\because \psi^* \text{ is optimal} \right ) 
    \end{split}
\end{equation*}
Taking square root we get\\ \begin{center}
    $\left \|-\nabla_\theta \mathbb{E}_{x\sim \mathcal{P}_X}\left [ g_{\psi}\left ( f_\theta \left ( x \right ) \right ) \right ]\right \| \leq \delta M\text{ which finishes the proof. }\square
$
\end{center}

\textbf{Theorem 3.2} indicates that the expected gradient of purely adversarial generator is proportional to spectral norm of Jacobian matrix and convergence error of discriminator. To put more succinctly, given a generator, the convergence error $\delta \rightarrow 0$ for a sufficiently trained discriminator. Thus, the adversarial discriminator does not produce erroneous gradients in the near optimal region, suggesting well behaved empirical risk.

\textbf{Augmented Objective:} Unlike sole supervision, the mapping function $f_\theta(.)$ in augmented objective has access to feedback signal from the discriminator. The optimization carried out in supervised learning with adversarial features is given by,
\begin{equation*}
\arg \min_{\theta} \mathbb{E}_{(x,y)\sim \mathcal{P}}\left [ l\left ( f_\theta(x);y \right ) - g_\psi \left ( f_\theta\left ( x \right ) \right ) \right ].
\end{equation*}
The discriminator cost function remains identical to Wasserstein discriminator as given by equation~(\ref{dis}).

\textbf{Theorem 3.3} \textit{Assume conditions of \textbf{Theorem 3.1} and \textbf{Theorem 3.2} hold. The expected gradient of supervised learning augmented with adversarial discriminator is bounded above by $(\lambda+\delta)M$. That is,
\begin{center}
$\left \| \nabla_\theta  \mathbb{E}_{(x,y)\sim \mathcal{P}}\left [ l\left ( f_\theta(x);y \right ) - g_\psi \left ( f_\theta\left ( x \right ) \right ) \right ] \right \| \leq \left ( \lambda+\delta \right )M$.
\end{center}
}

\textit{Proof.}
\begin{equation*}
\begin{split}
 \left \| \nabla_\theta  \mathbb{E}_{(x,y)\sim \mathcal{P}}\left [ l\left ( f_\theta(x);y \right ) - g_\psi \left ( f_\theta\left ( x \right ) \right ) \right ] \right \| & \leq \left \| \nabla_\theta  \mathbb{E}_{(x,y)\sim \mathcal{P}}\left [ l\left ( f_\theta(x);y \right )\right ] \right \| + \left \| - \nabla_\theta \mathbb{E}_{(x,y)\sim \mathcal{P}}\left [ g_\psi \left ( f_\theta\left ( x \right ) \right ) \right ] \right \| \\
& \leq \lambda M + \delta M~(\text{\textbf{Theorem 4.1} and \textbf{Theorem 4.2}})
\end{split}
\end{equation*}
So, we get
$\left \| \nabla_\theta  \mathbb{E}_{(x,y)\sim \mathcal{P}}\left [ l\left ( f_\theta(x);y \right ) - g_\psi \left ( f_\theta\left ( x \right ) \right ) \right ] \right \| \leq \left ( \lambda+\delta \right )M$   which finishes the proof. \hfill$\square$

According to \textbf{Theorem 3.3}, the expected gradient in augmented adversarial learning does not vanish in the near optimal region, i.e., $\left \| \Delta \theta\right \| \rightarrow \delta M$ as $\lambda \rightarrow 0$. In addition, the mapping function is guided through useful gradients from discriminator that allows efficient parametric update. Furthermore, the upper bound of \textbf{Theorem 3.2} ensures that mapping function, $f_\theta(.)$ remains within a small neighbourhood of optimal function approximator, $f_{\theta^*}(.)$.

\section{Supervised learning with adversarial features can be better than sole supervision}
\label{th45}
\subsection{Expected Empirical Risks}
\label{th4}
\textbf{Definition 4.1} We define the empirical risk in supervised learning with adversarial features and sole supervision as following.
\begin{equation*}
\begin{split}
\mathscr{R}_{aug} & := \inf_{\theta_N} \left \{ \mathbb{E}_{(x,y)\sim \mathcal{P}} \left [ \left \| l\left ( f_{\theta_N}^{aug}(x);y \right ) \right \| \right ] \right \} \\
\mathscr{R}_{sup} & := \inf_{\theta_N} \left \{ \mathbb{E}_{(x,y)\sim \mathcal{P}} \left [ \left \| l\left ( f_{\theta_N}^{sup}(x);y \right ) \right \| \right ] \right \}
\end{split}
\end{equation*}
Note that the total number of iterations ($N$) remains unchanged in both approaches. To compare supervised and adversarial learning, it is required that both methods be initialized with same set of parameters $\theta_0$.

\textbf{Theorem 4.1} \textit{Suppose the conditions of \textbf{Theorem 3.1} and \textbf{Theorem 3.2} holds. For a fixed number of iteration, the expected empirical risk of supervised learning augmented with adversarial features can be better than purely supervised learning. That is,
\begin{center}
$\mathscr{R}_{aug} \leq \mathscr{R}_{sup}$.
\end{center}
}

\textit{Proof.} Let both algorithms are initialized with $\theta_0$. From \textbf{Theorem 3.1} and \textbf{Theorem 3.3} we get,
\begin{center}
    $\left \|-\nabla_\theta \mathbb{E}_{x\sim \mathcal{P}_X}\left [ g_{\psi}\left ( f_\theta \left ( x \right ) \right ) \right ]\right \| \leq \delta M$ and
\end{center}

\begin{center}
$\left \| \nabla_\theta  \mathbb{E}_{(x,y)\sim \mathcal{P}}\left [ l\left ( f_\theta(x);y \right ) - g_\psi \left ( f_\theta\left ( x \right ) \right ) \right ] \right \| \leq \left ( \lambda+\delta \right )M$.
\end{center}

Since $f_\theta(x)$ is differentiable, each parametric update is undertaken in the following manner. 
\begin{equation*}
    \begin{split}
        f_{\theta_n}(x) - f_{\theta_{n+1}}(x) & = \nabla_\theta f_\theta(x)\rvert_{\theta_n}(\theta_n - \theta_{n+1}), n=0,1,\dots,N-1\\
        \implies f_{\theta_0}(x) - f_{\theta_{N}}(x) & = \displaystyle\sum\limits_{i=0}^{N-1} \nabla_\theta f_\theta(x)\rvert_{\theta_i}(\theta_i - \theta_{i+1})
    \end{split}
\end{equation*}
 Now, 
 \begin{equation*}
     \begin{split}
         \mathbb{E}_{(x,y)\sim\mathcal{P}}[\lVert f_{\theta_0}(x) - f_{\theta_{N}}^{aug}(x) \rVert] & \leq  \mathbb{E}_{(x,y)\sim\mathcal{P}}[ \displaystyle\sum\limits_{i=0}^{N-1} \lVert \nabla_\theta f_\theta(x)\rvert_{\theta_i}\rVert \lVert(\theta_i - \theta_{i+1}) \rVert]\\
         & \leq M  \displaystyle\sum\limits_{i=0}^{N-1} \lVert(\theta_i - \theta_{i+1})\rVert,~(\because \mathbb{E}_{(x,y)\sim \mathcal{P}}\left [ \left \| \mathcal{J}_{\theta}\left ( f_{\theta}\left ( x \right ) \right ) \right \| \right ]\leq M)\\
         & \leq  M \displaystyle\sum\limits_{i=0}^{N-1} \left \| \nabla_\theta  \mathbb{E}_{(x,y)\sim \mathcal{P}}\left [ l\left ( f_\theta(x);y \right ) - g_\psi \left ( f_\theta\left ( x \right ) \right ) \right ]|_{\theta_i} \right \| \\
         & \leq  MN \sup_{i\in [0,N-1]} \left \{  \left \| \nabla_\theta  \mathbb{E}_{(x,y)\sim \mathcal{P}}\left [ l\left ( f_\theta(x);y \right ) - g_\psi \left ( f_\theta\left ( x \right ) \right ) \right ]|_{\theta_i} \right \| \right \} \\
         & \leq  M^2N(\lambda+\delta),~( \text{from \textbf{Theorem 3.3}})
     \end{split}
 \end{equation*}
Similarly, for purely supervised learning we get \begin{center}
    $ \mathbb{E}_{(x,y)\sim\mathcal{P}}[\lVert f_{\theta_0}(x) - f_{\theta_{N}}^{sup}(x) \rVert ]\leq M^2N\lambda$.
\end{center}

By locally $K-$Lipschitz continuous property of loss function,
\begin{equation*}
    \begin{split}
        \mathbb{E}_{(x,y)\sim\mathcal{P}}[\lVert l\left ( f_{\theta_0}(x);y \right ) - l\left ( f_{\theta_N}^{aug} (x);y \right ) \rVert] & \leq K \mathbb{E}_{(x,y)\sim\mathcal{P}}[\lVert f_{\theta_0}(x) - f_{\theta_{N}}^{aug}(x) \rVert] \\
        & \leq KM^2N(\lambda+\delta).
    \end{split}
\end{equation*}
Similarly, 
\begin{equation*}
    \begin{split}
        \mathbb{E}_{(x,y)\sim\mathcal{P}}[\lVert l\left ( f_{\theta_0}(x);y \right ) - l\left ( f_{\theta_N}^{sup} (x);y \right ) \rVert] & \leq K \mathbb{E}_{(x,y)\sim\mathcal{P}}[\lVert f_{\theta_0}(x) - f_{\theta_{N}}^{sup}(x) \rVert] \\
        & \leq KM^2N\lambda\\
        \implies \mathbb{E}_{(x,y)\sim\mathcal{P}}[\lVert l\left ( f_{\theta_0}(x);y \right )\rVert] - \mathbb{E}_{(x,y)\sim\mathcal{P}}[\lVert l\left ( f_{\theta_N}^{sup} (x);y \right ) \rVert]  &\leq KM^2N\lambda\\
        \implies \mathbb{E}_{(x,y)\sim\mathcal{P}}[\lVert l\left ( f_{\theta_N}^{sup} (x);y \right ) \rVert] & \geq \mathbb{E}_{(x,y)\sim\mathcal{P}}[\lVert l\left ( f_{\theta_0}(x);y \right )\rVert] -  KM^2N\lambda\\
    \end{split}
\end{equation*}

For augmented objective, 
\begin{equation*}
    \begin{split}
        \mathbb{E}_{(x,y)\sim\mathcal{P}}[\lVert l\left ( f_{\theta_N}^{aug} (x);y \right ) \rVert] & \geq \mathbb{E}_{(x,y)\sim\mathcal{P}}[\lVert l\left ( f_{\theta_0}(x);y \right )\rVert] -  KM^2N(\lambda+\delta)\\
        &\geq \mathbb{E}_{(x,y)\sim\mathcal{P}}[\lVert l\left ( f_{\theta_0}(x);y \right )\rVert] - KM^2N\lambda - KM^2N\delta\\
        & \geq \inf_{\theta_N}\left \{ \mathbb{E}_{(x,y)\sim\mathcal{P}}[\lVert l\left ( f_{\theta_N}^{sup} (x);y \right ) \rVert] \right \} - KM^2N\delta
    \end{split}
\end{equation*}
Thus, the expected empirical risk of supervised learning with adversarial features becomes,
\begin{equation*}
    \begin{split}
        \inf_{\theta_N}\left \{ \mathbb{E}_{(x,y)\sim\mathcal{P}}[\lVert l\left ( f_{\theta_N}^{aug} (x);y \right ) \rVert] \right \} & = \inf_{\theta_N}\left \{ \mathbb{E}_{(x,y)\sim\mathcal{P}}[\lVert l\left ( f_{\theta_N}^{sup} (x);y \right ) \rVert] \right \} - KM^2N\delta \\
        \implies \mathscr{R}_{aug} & = \mathscr{R}_{sup} - KM^2N\delta \\
        \implies \mathscr{R}_{aug} &\leq \mathscr{R}_{sup},~(\because KM^2N\delta \geq=0).\text{ This finishes the proof.}~\square
    \end{split}
\end{equation*}

\subsection{Rate of Convergence}
\label{th5}
\textbf{Definition 4.2} We define $N^*$ to be the minimum number of iterations required to achieve optimal set of parameters, provided it exists.

\textbf{Theorem 4.2} \textit{If the conditions of \textbf{Theorem 3.1} and \textbf{Theorem 3.2} are satisfied, then it is cheaper to achieve optimal set of parameters in case of augmented objective as compared to sole supervision. That is,
\begin{center}
$ N_{aug}^* \leq N_{sup}^*$.
\end{center}
}

\textit{Proof.} Let $N$ denote the number of iterations required to attain optimal set of parameters $\theta^*$. The trainable parameters are updated as per the following rule.
\begin{equation*}
    \begin{split}
        &  \theta_{n+1} = \theta_n - \Delta\theta_n,~n=0,1,\dots,N-1\\
         \implies &\theta^* = \theta_0 - \sum_{i=0}^{N-1} \Delta\theta_i \\
        \implies &\lVert \theta^* - \theta_0 \rVert  =  \lVert - \sum_{i=0}^{N-1} \Delta\theta_i \rVert \leq  \sum_{i=0}^{N-1} \lVert \Delta\theta_i \rVert \leq N \sup_{i\in [0,N-1]}\left \{ \lVert \Delta\theta_i\rVert \right \}\\
        \implies & N \geq \frac{ \lVert \theta^* - \theta_0 \rVert}{\sup_{i\in [0,N-1]}\left \{ \lVert \Delta\theta_i\rVert \right \}} 
    \end{split}
\end{equation*}
Therefore, the minimum number of iterations required to achieve optimal empirical risk is given by,
\begin{equation*}
    \begin{split}
        N_{aug}^* & = \frac{ \lVert \theta^* - \theta_0 \rVert}{\left ( \lambda+\delta \right )M}~\text{(\textbf{ from Theorem 3.3})}\\
        N_{sup}^* & = \frac{ \lVert \theta^* - \theta_0 \rVert}{\lambda M.}~\text{(\textbf{ from Theorem 3.1})}
    \end{split}
\end{equation*}
Now, taking ratio of both optimal iterations,
\begin{equation*}
    \begin{split}
        & \frac{N_{aug}^*}{N_{sup}^*}  = \frac{\lambda M}{\left ( \lambda+\delta\right)M} \\
        \implies & N_{aug}^* = N_{sup}^*\left ( \frac{\lambda}{ \lambda+\delta} \right ) \\
        \implies & N_{aug}^* \leq N_{sup}^*,~(\because \delta \geq 0).~\text{This finishes the proof.}~\square
    \end{split}
\end{equation*}

\section{Conclusion}
\label{conc}
In this study, we investigated the reason behind slow convergence of purely supervised learning in near optimal region. Further, our analysis showed how adversarial features contribute towards mitigating this convergence issue. Finally, we provided theoretical proofs to corroborate our main result that shows supervised learning with adversarial regularization can be better than purely supervised learning. At the end, we complemented our hypothesis by showing that the expected empirical risk and rate of convergence is relatively better in regularized adversarial learning as compared to sole supervision.

	\ifCLASSOPTIONcaptionsoff
	\newpage
	\fi
	\bibliographystyle{ieeetr}
	\bibliography{egbib}

\end{document}